\documentclass[journal]{IEEEtran}

\ifCLASSINFOpdf
\else
\fi

\usepackage{times}
\usepackage{epsfig}
\usepackage{graphicx}
\usepackage{amsmath}
\usepackage{amssymb}
\usepackage{dsfont}
\usepackage{multirow}
\usepackage{epstopdf}
\usepackage{bmpsize}
\usepackage{enumerate}
\usepackage{hyperref}
\usepackage{array}
\usepackage{color}
\usepackage{multirow}
\usepackage{amsmath}
\usepackage{cite}
\usepackage[misc]{ifsym}
\usepackage{subfig}
\usepackage{authblk}

\begin{document}
%
\title{Activation to Saliency: Forming High-Quality Labels for Completely Unsupervised Salient Object Detection}
%
%
%

\author{Huajun~Zhou,~\IEEEmembership{Graduate~Student~Member,~IEEE,}
        Peijia~Chen,
        Lingxiao~Yang,~\IEEEmembership{Member,~IEEE,}
        Jianhuang~Lai,~\IEEEmembership{Senior~Member,~IEEE,}
        Xiaohua~Xie,~\IEEEmembership{Member,~IEEE.} 
\thanks{Manuscript received XXX XX, XXXX; revised XXX XX, XXXX. This work was supported in part by the National Natural Science Foundation of China under Grant 62072482, the Key-Area Research and Development Program of Guangdong Province under Grant 2019B010155003, and the Guangdong-Hong Kong-Macao Greater Bay Area International Science and Technology Innovation Cooperation Project under Grant 2021A0505030080. (Corresponding author: Xiaohua Xie.)}
\thanks{The authors are with the School of Computer Science and Engineering, the Guangdong Province Key Laboratory of Information Security Technology, and the Key Laboratory of Machine Intelligence and Advanced Computing, Ministry of Education, Sun Yat-sen University, Guangzhou 510006, China (e-mail: zhouhj26@mail2.sysu.edu.cn; chenpj8@mail2.sysu.edu.cn; yanglx9@mail.sysu.edu.cn; stsljh@mail.sysu.edu.cn; xiexiaoh6@mail.sysu.edu.cn).}}




%
%

\markboth{IEEE TRANSACTIONS ON XXX XXX}{}
%



\maketitle

\begin{abstract}
Existing deep learning-based Unsupervised Salient Object Detection (USOD) methods rely on supervised pre-trained deep models. Moreover, they generate pseudo labels based on hand-crafted features, which lack high-level semantic information. In order to overcome these shortcomings, we propose a new two-stage Activation-to-Saliency (A2S) framework that effectively excavates high-quality saliency cues to train a robust saliency detector. It is worth noting that our method does not require any manual annotation even in the pre-training phase. In the first stage, we transform an unsupervisedly pre-trained network to aggregate multi-level features to a single activation map, where an Adaptive Decision Boundary (ADB) is proposed to assist the training of the transformed network. Moreover, a new loss function is proposed to facilitate the generation of high-quality pseudo labels. In the second stage, a self-rectification learning paradigm strategy is developed to train a saliency detector and refine the pseudo labels online. In addition, we construct a lightweight saliency detector using two Residual Attention Modules (RAMs) to largely reduce the risk of overfitting. Extensive experiments on several SOD benchmarks prove that our framework reports significant performance compared with existing USOD methods. Moreover, training our framework on 3,000 images consumes about 1 hour, which is over 30$\times$ faster than previous state-of-the-art methods.
Code will be published at \url{https://github.com/moothes/A2S-USOD}.
\end{abstract}


\section{Introduction}
\noindent
\IEEEPARstart{R}{}esearches on supervised Salient Object Detection (SOD) have reached impressive achievements \cite{f3net, inv, rfcn, ucf, page, gl, dcl, dscn, scrn} owing to the developments of Convolutional Neural Networks (CNNs) \cite{ghost, res2net, efficientnet, mobilev2}.
An essential prerequisite for these advancements is the large-scale high-quality human-labeled datasets.
However, the pixel-level annotations for salient objects is laborious.
Therefore, Unsupervised Salient Object Detection (USOD) receives increasing attention because it does not require extra efforts for annotating.
The main challenges of USOD are how to model the image saliency with prior knowledge and generate high-quality pseudo labels for training a saliency detector.

\begin{figure}[!t]
\includegraphics[width=3.3in]{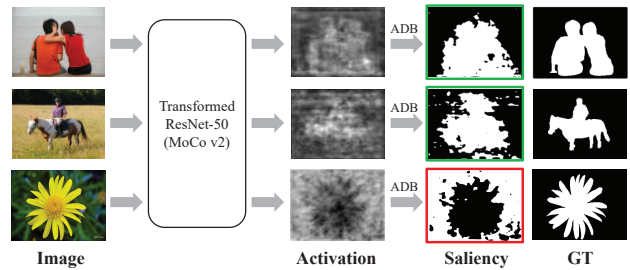}
\caption{Saliency information extracted from the activation map $\boldsymbol M$ using the proposed adaptive decision boundary (ADB).
Green and red blocks mean correct and inversed saliency maps, respectively.}
\label{fig:framework}
\end{figure}

Traditional SOD methods based on hand-crafted features can well segment some regions with conspicuous colors, but struggle to capture more complex salient objects, because hand-crafted features lack high-level semantic information.
Existing deep learning-based USOD methods \cite{sbf, mnl, usps} fuse saliency predictions by multiple traditional SOD methods as saliency cues, and refine them with the assistant of semantic information.
The semantic information is obtained from the models trained by supervised learning with other vision tasks, such as object recognition \cite{vgg, resnet} and semantic segmentation \cite{deeplab, cityscape}.
Methods in \cite{sbf, mnl} directly use these saliency cues as pseudo labels to train a saliency detector with the help of some auxiliary models such as fusion and noise models. Recent USPS method\cite{usps} utilizes these saliency cues to train multiple networks and refine them using intermediate predictions.
However, hand-crafted feature-based methods usually generate low-confidence regions, which greatly limits the effectiveness of existing USOD methods.

\begin{figure*}[!t]
\includegraphics[width=7.2in,height=1.05in]{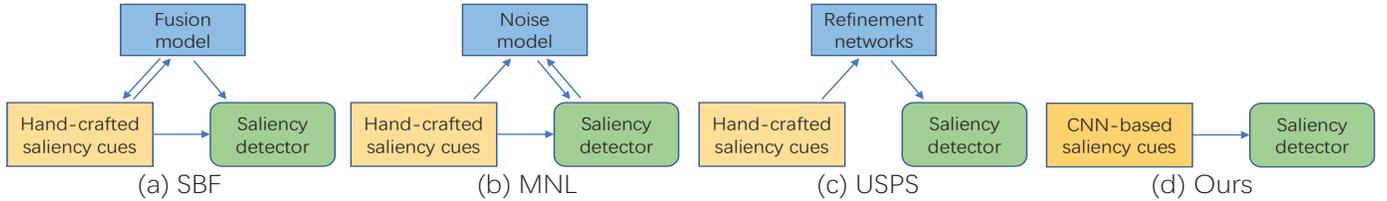}
\caption{Pipelines for deep learning-based USOD methods. Existing methods contain three stages, which are denoted by different colors. Yellow: extracting saliency cues; Blue: constructing an auxiliary model; Green: training a saliency detector. Since our framework extracts high-quality saliency cues, it only contains two stages to tackle the USOD task.}
\label{fig:pipeline}
\end{figure*}

These years, multiple researchers \cite{yosinski, zhaodiversified, xiaoapp, mahendran} have made efforts to understand the features extracted by deep networks.
A few works \cite{cam, gradcam, rcnn, vis} have proved that, unlike hand-crafted features that are intuitive to the human vision system, CNNs pre-trained on large-scale data usually produce high activations on some primary objects, yet are difficult to activate other non-salient objects and background within an image.
It means that pre-trained networks are capable of differentiating salient objects and background in images.
Moreover, multi-level features extracted by pre-trained networks are rich in semantic information.
Therefore, they can help us generate more semantically reasonable pseudo labels than hand-crafted features.
To prove this point, we transform the ResNet-50 \cite{resnet} which is pre-trained by MoCo v2 \cite{mocov2} to aggregate multi-level feature maps as a single activation map, denoted as $\boldsymbol M$.
As shown in Fig. \ref{fig:framework}, some regions in $\boldsymbol  M$ are distinctive from other pixels.
By refining these activations, we can extract high-quality saliency cues to serve as pseudo labels for training a salient object detector.

In this work, we propose an efficient two-stage Activation-to-Saliency (A2S) framework for the Unsupervised Salient Object Detection (USOD) task.
The proposed A2S framework uses the features extracted by an unsupervisedly pre-trained network, instead of using hand-crafted features, to better localize salient objects.
In the first stage, multi-level features in a pre-trained network are integrated by four auxiliary Squeeze-and-Excitation (SE) blocks \cite{senet} to produce an activation map $\boldsymbol M$ for each image.
Based on our observation illustrated in Fig. \ref{fig:framework}, we propose to employ a pixel-wise linear classifier to find saliency cues.
Due to the diverse contents in different images, learning a single classifier for all images is sub-optimal.
Instead, we design an image-specific classifier and form the Adaptive Decision Boundary (ADB).
Based on ADB, we develop a loss function to enlarge the distances between features and their means.
Note that our linear classifier is trained with the proposed loss function without manual annotations.
In the second stage, we propose a self-rectification learning method to refine pseudo labels.
To achieve rectification, an Online Label Rectifying (OLR) strategy is proposed to reduce the negative impact of distractors by updating pseudo labels online.
To reduce the risk of overfitting, we construct a lightweight saliency detector using two novel Residual Attention Modules (RAMs).
The proposed RAM enhances the topmost encoder feature using low-level features.

Extensive experiments prove that the proposed framework achieves state-of-the-art performance against existing USOD methods and is competitive to some latest supervised SOD methods.
In addition, our framework consumes about 1 hour to train with 3000 images, which is about 30$\times$ faster than previous state-of-the-art USOD methods.

In summary, our main contributions are as follows.
\begin{itemize}
\item We propose an efficient Activation-to-Saliency (A2S) framework for completely Unsupervised Salient Object Detection (USOD).
\item We propose an Adaptive Decision Boundary (ADB) to facilitate the extraction of high-quality saliency cues from images.
\item We propose an Online Label Rectifying (OLR) strategy to reduce the negative effect of distractors during training.
\item We propose a lightweight saliency detector to largely reduce the risk of overfitting.
\end{itemize}

\begin{figure*}[t]
\centering
\includegraphics[width=7.2in,height=3.5in]{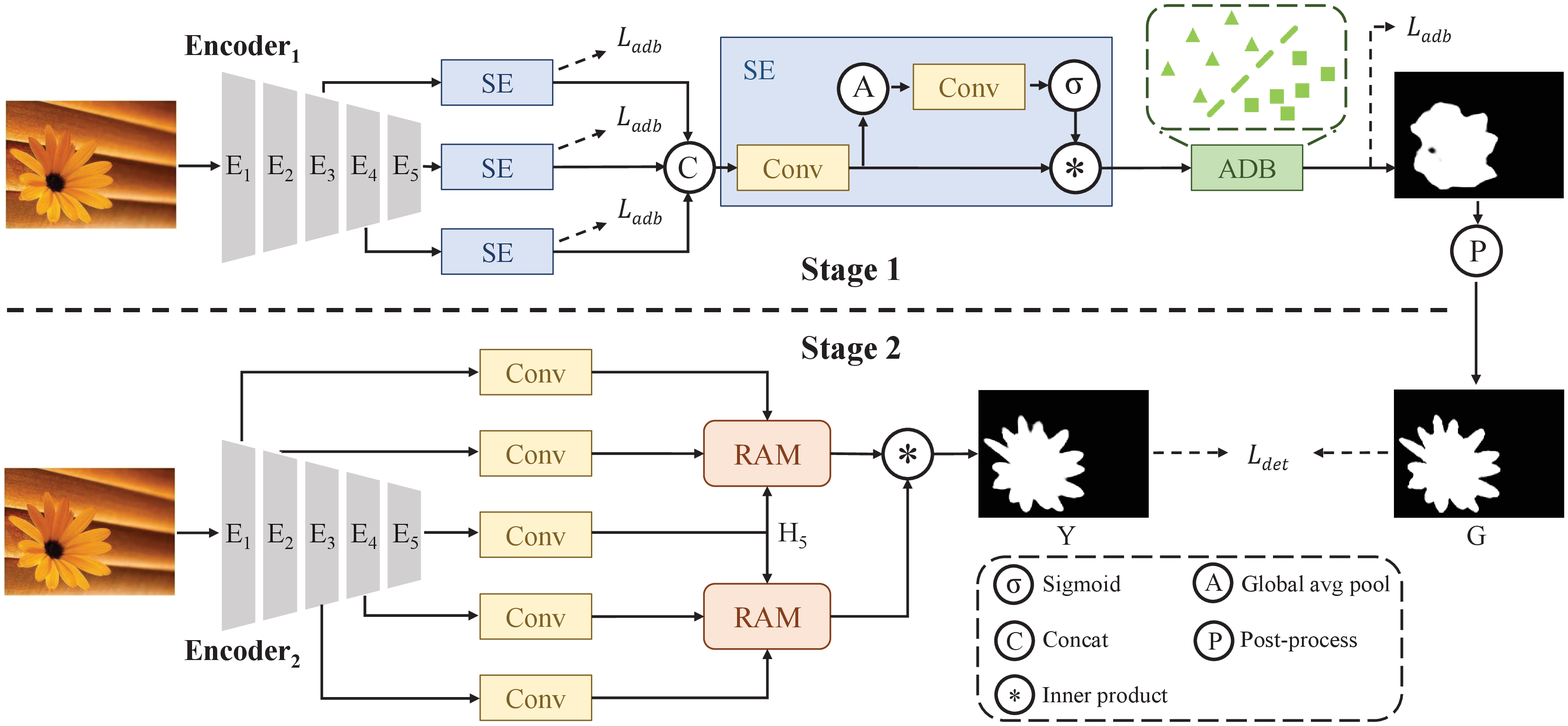}
\caption{Our framework with two stages, one for saliency cue extraction and another for self-rectification learning.
Two encoders of two stages are initialized by the same weights. Blocks with the same colors have the same structure, but have independent weights.}
\label{fig:sal}
\vspace{-0.2cm}
\end{figure*}

\section{Related Works}
\subsection{SOD by Unsupervised Hand-crafted Methods}
Conventional SOD methods \cite{dsr, mc, rbd, cssd} extracted saliency cues from images by using different priors and hand-crafted features.
Li et al. \cite{dsr} computed dense and sparse reconstruction based on the background templates for each image region, and predicted the pixel-level saliency with the integration of multi-scale reconstruction errors.
Jiang et al. \cite{mc} predicted the saliency maps by calculating the distances between boundary superpixels and non-boundary superpixels.
Zhu et al. \cite{rbd} proposed a robust background measure to characterize the spatial layout of image regions with respect to image boundaries, and employed an optimization framework to integrate multiple low-level information.
Yan et al. \cite{cssd} computed multiple saliency cues from three over-segmented maps and fed them into a tree-structure graphical model to get the final results.

These methods can segment conspicuous regions in images but fail to capture salient objects because hand-crafted features cannot well model semantic information.

\subsection{SOD by Supervised Deep-Learning}
In recent researches, most of supervised DL-based SOD methods \cite{amulet, dss, cag, contour, fcn, poolnet} improved their performances by enhancing a U-shape structure \cite{unet}.
Liu et al. \cite{picanet} proposed a pixel-wise contextual attention network to enhance the learned features using context information.
Luo et al. \cite{nldf} developed the contrast features that subtract each feature from its local average to enhance the features in skip connections.
Liu et al. \cite{dhsnet} used a Recurrent Convolution Layer (RCL) to hierarchically and progressively render image details.
Zhao et al. \cite{pfa} divided five encoder features into two branches, and aggregated the fused features in these branches to produce the final predictions.
Wang et al. \cite{srm} fused high-level semantic knowledge and spatially rich information of low-level features from two different encoders to produce more robust predictions.
Zhao et al. \cite{gate} designed a novel gated dual branch structure to build the cooperation among different levels of features.
In addition, \cite{pagrn, sac, bmp, mlm, PFPN} improved the performance by introducing various feature fusion modules.

Recent works noticed that edge information can assist SOD methods to produce more accurate boundary.
Feng et al. \cite{afnet} employed contour maps to supervise the edge of saliency predictions.
Li et al. \cite{ckt} proposed an alternative structure that saliency and contour maps are utilized as supervision signal for each other.
Zhao et al. \cite{egnet} supervised the most shallow encoder feature using edges generated from images and integrated it with other higher-level encoder features.
Zhou et al. \cite{itsd} constructed a two stream decoder that integrates contour and saliency information to each other alternatively.
Wei et al. \cite{ldf} decoupled the salient objects into body and contour maps, and employed three decoders to output these maps as well as saliency maps, respectively.

Although these methods achieved impressive results on SOD benchmarks, they require a significant amount of human-labeled data for training, which are expensive to collect.

\subsection{SOD by Unsupervised Deep-Learning}
A few works \cite{sbf, mnl, usps} attempted to tackle USOD task via Deep-Learning techniques.
They learned saliency from multiple noisy saliency cues produced by traditional SOD methods, as shown in Fig. \ref{fig:pipeline}.
Next, they refine these saliency cues using semantic information from some supervised methods in other related vision tasks, such as object recognition \cite{vgg, resnet} and semantic segmentation \cite{deeplab, cityscape}.
Zhang et al. \cite{sbf} learned saliency by using the intra-image fusion stream and inter-image fusion stream to produce multi-level weights for the noisy saliency cues.
Zhang et al. \cite{mnl} modeled the noise of each pixel as a zero-mean Gaussian distribution and reconstructed noisy saliency cues by integrating saliency predictions and randomly sampled noise.
Recently, Nguyen et al. \cite{usps} claimed that directly using these noisy saliency cues to train a saliency detector is sub-optimal.
Therefore, they first train multiple refinement networks to improve the quality of these saliency cues.
Subsequently, they generated multiple homologous saliency maps by excavating inter-image consistency between these cues to train a saliency detector.
Although it has reported impressive results, training a series of refinement networks greatly reduces its efficiency.

Unlike previous DL-based USOD methods still benefit from some supervised methods, all components in our framework are trained in a completely unsupervised manner, including encoder (MoCo v2 \cite{mocov2}), decoder, and saliency detector.
Moreover, instead of extracting noisy saliency cues using traditional SOD methods, we present a novel perspective to excavate high-quality saliency cues based on the learned features of a pre-trained network.
Our saliency cues are likely to capture salient objects because high-level features in the encoder have concluded rich semantic information.
Using our high-quality saliency cues as pseudo labels, we can train robust saliency detectors without the assistant of any auxiliary model.

\begin{figure}[!t]
\centering
\begin{minipage}{1 \textwidth}
\subfloat[Image]{
\includegraphics[width=0.82in,height=0.78in]{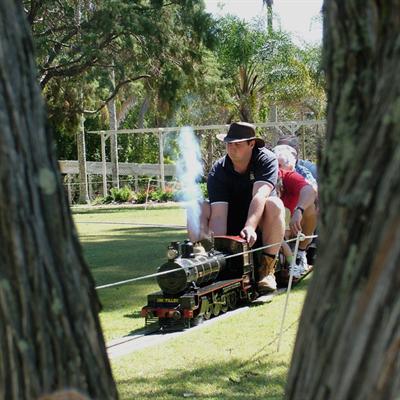}}
\subfloat[GT]{
\includegraphics[width=0.82in,height=0.78in]{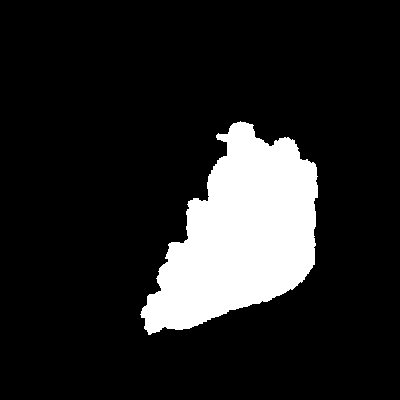}}
\subfloat[Activation]{
\includegraphics[width=0.82in,height=0.78in]{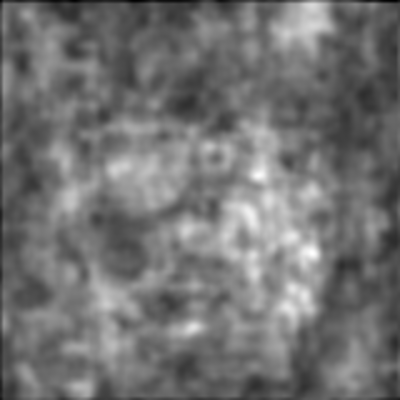}}
\subfloat[Ours]{
\includegraphics[width=0.82in,height=0.78in]{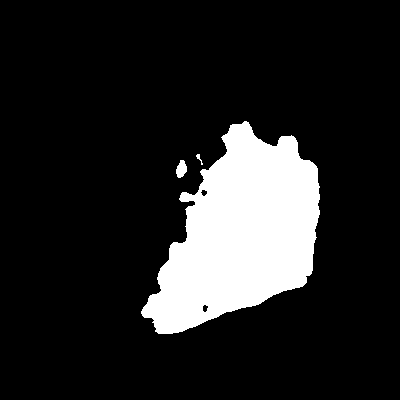}}
\end{minipage}

\vspace{-0.3cm}
\begin{minipage}{1 \textwidth}
\subfloat[DSR \cite{dsr}]{
\includegraphics[width=0.82in,height=0.78in]{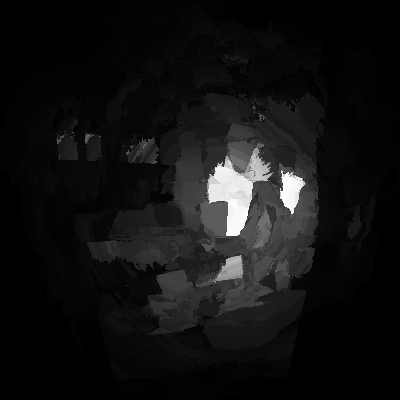}}
\subfloat[MC \cite{mc}]{
\includegraphics[width=0.82in,height=0.78in]{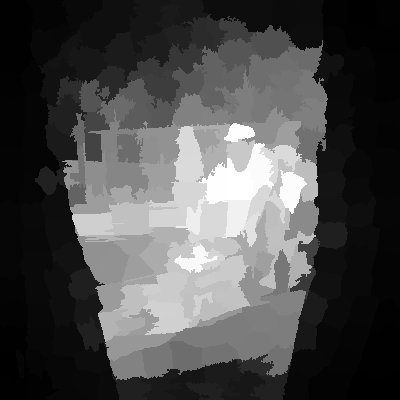}}
\subfloat[RBD \cite{rbd}]{
\includegraphics[width=0.82in,height=0.78in]{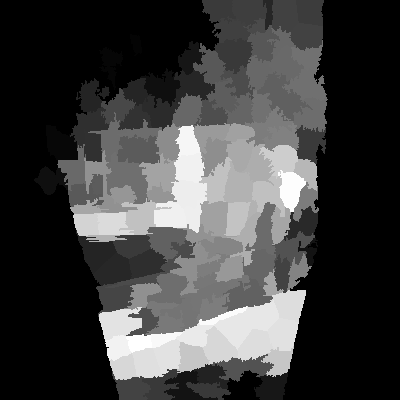}}
\subfloat[HS \cite{cssd}]{
\includegraphics[width=0.82in,height=0.78in]{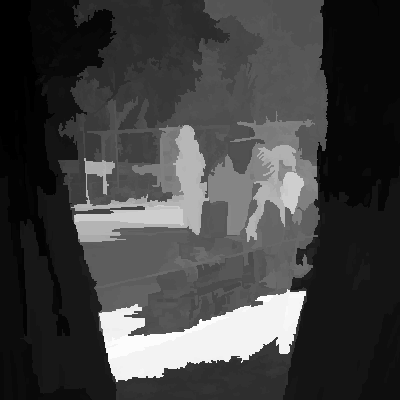}}
\end{minipage}

\caption{Examples of the extracted saliency cues.}
\label{fig:pseudo}
\end{figure}

\section{Proposed Approach}
In this section, we elaborate details about the proposed A2S framework, as shown in Fig. \ref{fig:sal}.
Our framework contains two stages, including a stage for saliency cue extraction and a stage for self-rectification learning.

\subsection{Stage 1: Saliency Cue Extraction}
\textbf{Inconsistency between Appearance and Semantic.}
The goal of USOD task is to segment the whole salient objects in images instead of some conspicuous regions.
However, traditional SOD methods are prone to segmenting regions based on their appearances.
Due to the inconsistency between region appearances and object semantics, these methods are unlikely to capture salient objects in some challenging images.
For example, in Fig. \ref{fig:pseudo}, DSR \cite{dsr} (e) focuses on partial salient objects.
MC \cite{mc} (f) captures salient objects but fails to differentiate such objects with their surroundings.
RBD \cite{rbd} and HS \cite{hs} (g-h) detect regions with distinctive colors.
In activation map (c), we observe high activations around people regions.
This activation map can be utilized to generate high-quality saliency cues (d).

\textbf{Activation Synthesis.}
Our method is based on the pre-trained deep network to generate high-quality pseudo labels.
Given an unsupervisedly pre-trained network (e.g., MoCo v2 \cite{mocov2}), high-level features often contain more semantic information but lose many details. 
Low-level features usually have more activations on texture details but fail to capture global statistics.
Therefore, we leverage multi-level features in the proposed framework.
Due to the lack of manual annotations, training too much parameters greatly increases the risk of overfitting.
Moreover, it may destroy semantic information learned in the pre-trained network and cause the network hard to converge, as demonstrated in our experiments.
Thus, we add several auxiliary blocks to the pre-trained network and only train those blocks.
Specifically, outputs of stages 3, 4, 5 from the pre-trained network are denoted as $\boldsymbol E_3$, $\boldsymbol E_4$, and $\boldsymbol E_5$, respectively.
Each feature $\boldsymbol E_i$ is processed by a Squeeze-and-Excitation (SE) block to enhance the learned representations, denoted as $\boldsymbol F_i$.
Another SE block is employed to generate the fused feature map $\boldsymbol F$ by integrating $\boldsymbol F_3$, $\boldsymbol F_4$, and $\boldsymbol F_5$.
We define the above procedures as our \textbf{transformed network}:
\begin{equation}
\label{eqn:s1}
\begin{split}
\boldsymbol E_3, \boldsymbol E_4, \boldsymbol E_5 &= Encoder_{1}(\boldsymbol X),\\
\boldsymbol F_i &= SE_i(\boldsymbol E_i), i=3,4,5\\
\boldsymbol F &= SE_0(concat(\boldsymbol F_3, \boldsymbol F_4, \boldsymbol F_5)),
\end{split}
\end{equation}
where $\boldsymbol X$ is input image.
We simplify the transformed network as $\boldsymbol F = \phi_{act}(\boldsymbol X)$ for the following illustration.

\begin{figure}[!t]
\centering
\begin{minipage}{1 \textwidth}
\subfloat[Image]{
\includegraphics[width=1.1in,height=0.78in]{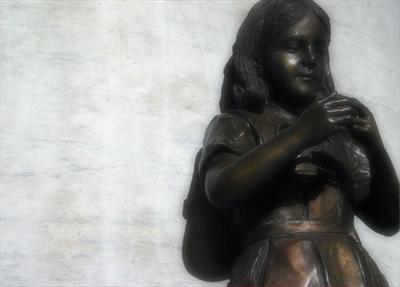}}
\subfloat[GT]{
\includegraphics[width=1.1in,height=0.78in]{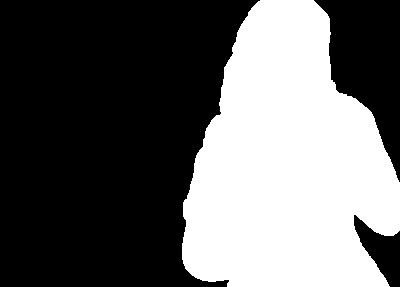}}
\subfloat[Activation]{
\includegraphics[width=1.1in,height=0.78in]{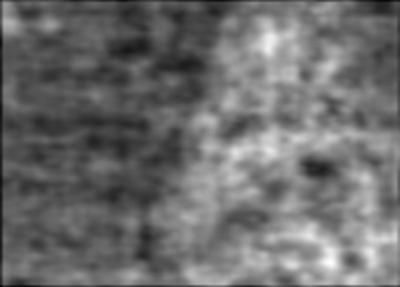}}
\end{minipage}

\vspace{-0.3cm}
\begin{minipage}{1 \textwidth}
\subfloat[Otsu \cite{otsu}]{
\includegraphics[width=1.1in,height=0.78in]{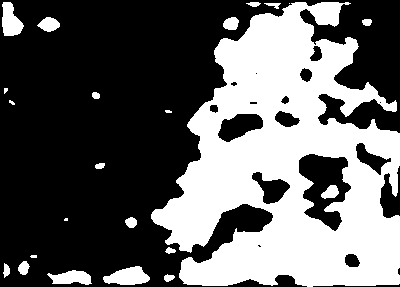}}
\subfloat[Mean]{
\includegraphics[width=1.1in,height=0.78in]{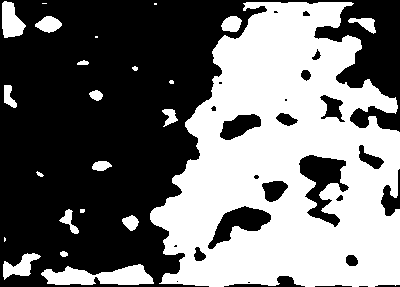}}
\subfloat[Median]{
\includegraphics[width=1.1in,height=0.78in]{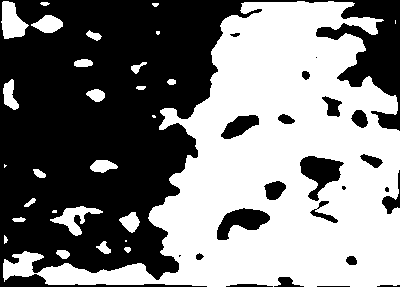}}
\end{minipage}

\caption{Saliency cues generated by different thresholds before training our transformed network.}
\label{fig:cues}
\end{figure}

\begin{table}[t]
\begin{center}
\caption{Inter-class variance for different strategies.}
\renewcommand\arraystretch{1.2} 
\label{tab:otsu}
\begin{tabular}{c|ccc}
\hline
Strategy    & Otsu & Mean & Median \\\hline
Variance & 0.1408 & 0.1405 & 0.1365  \\ \hline
\end{tabular}
\end{center}
\end{table}

\textbf{Adaptive Decision Boundary.}
Given the feature set $\boldsymbol F = \{ \boldsymbol f^{n}\} = \{\phi_{act}(\boldsymbol x^{n})\}$, where $n$ indicates the spatial index of feature.
$\boldsymbol F$ contains multiple feature maps that are stimulated by different conspicuous regions.
To gather these regions in each image, we sum the feature maps in $\boldsymbol F$ to generate a single activation map $\boldsymbol M$.
In this way, each $\boldsymbol f^{n}$ is converted to a scalar $\boldsymbol 1^{T} \boldsymbol f^n$.
As shown in Fig. \ref{fig:framework}, $\boldsymbol M$ reveals the potential saliency regions in each image.
A coarse saliency prediction can be generated by binarizing $\boldsymbol M$ using a threshold.
There are various strategies to produce a threshold for each image, such as Otsu algorithm \cite{otsu}, mean and median value of $\boldsymbol M$.
For each image, we split all pixels into two groups using these strategies.
As recommended by \cite{otsu}, we consider that larger inter-group variance $V$ indicates a better threshold.
The inter-group variance $V$ can be computed by:
\begin{equation}
   V = p_{1}p_{2}(\mu_{1} - \mu_{2})^2,
\end{equation}
where $p$ and $\mu$ indicate the probability and average value of each group, respectively.
Finally, the final score of each strategy is obtained by averaging the corresponding $V$ across train and validation subsets of MSRA-B.
As we can see in Tab. \ref{tab:otsu}, Otsu algorithm reports the maximum $V$ value compared to other strategies, which means that it can compute a better threshold to binarize $\boldsymbol M$.
However, this strategy is relatively slow because it uses an iterative process to find the optimal threshold for each image.
Interestingly, we find that $\boldsymbol 1^{T} \bar{\boldsymbol f}$, mean value of $\{\boldsymbol 1^{T} \boldsymbol f^n\}$, reaches a similar result as Otsu algorithm.
As shown in Fig. \ref{fig:cues}, $\boldsymbol 1^{T} \bar{\boldsymbol f}$ is slightly inferior than Otsu algorithm but apparently much more efficient.
Thus, we propose to use image-specific mean values to binarize $\boldsymbol M$.

In the activation map $\boldsymbol M$, $\boldsymbol 1^{T} \bar{\boldsymbol f}$ is selected as threshold to split all pixels into two different groups.
Meanwhile, pixel with large distance to $\boldsymbol 1^{T} \bar{\boldsymbol f}$ indicates that it is easy to be distinguished.
This process is equivalent to finding a decision value $\theta ^{n}$ of $n$-th pixel on linear decision boundary:
\begin{equation}
\label{eqn:db}
   \theta ^{n} = \boldsymbol 1^{T}\boldsymbol f^{n} - \boldsymbol 1^{T}\bar{\boldsymbol f} = \boldsymbol 1^{T}(\boldsymbol f^{n} - \frac{1}{N}\sum_{n}^{N}{\boldsymbol f^{n}}),
\end{equation}
where $N$ is the number of features in $\boldsymbol F$.
Since $\boldsymbol 1^{T} \bar{\boldsymbol f}$ is adaptive to images, Eqn. \ref{eqn:db} is an Adaptive Decision Boundary (ADB) for each image.

Different from adjusting decision boundaries to fit the fixed features, parameters in our decision boundary are adaptive to images.
The most simple method is directly using $\theta ^{n}$ as decision value to generate pseudo labels.
However, we expect that features $\boldsymbol f^{n}$ have larger distances to $\bar{\boldsymbol f}$ to make them become more distinctive.
Thus, we propose the following loss function:
\begin{equation}
\begin{split}
    D_n &= |\sigma(\theta ^{n}) - \sigma(\theta ^{\ast})| \\
    L_{adb} &= - \frac{1}{N}\sum^{N}_{n}({\|D_n\|_2 + \alpha \|D_n\|_1})
\end{split}
\label{eqn:loss_1}
\end{equation}
where $\sigma$ is sigmoid function and $\alpha$ is a hyperparameter.
$\|\cdot\|_1$ and $\|\cdot\|_2$ are L1 and L2 distance, respectively.
$\theta ^{\ast}$ is set to 0 because it indicates the decision values of pixels on decision boundary.
By maximizing distance $D_n$, the network is trained to extract more distinctive features based on our adaptive decision boundary.
The L2 distance promotes the network to focus on distinctive samples far away from ADB, meanwhile, the L1 distance prevents the gradients of hard examples from vanishing.
$F_3$, $F_4$, $F_5$ and $F$ are supervised by this loss to promote the network to learn more robust representations.
For each iteration, half of pixels are randomly dropped in the loss function to alleviate the overfitting problem.

Although the above methods can locate salient regions based on activation map $\boldsymbol M$, their predictions exist some issues.
Firstly, as shown in Fig. \ref{fig:framework}, our network may output inversed saliency maps for images.
In another word, given two sets of features: $S_{0} = \{\boldsymbol f^{n}, where \ \theta ^{n} < 0\}$ and $S_{1} = \{\boldsymbol f^{n}, where \  \theta ^{n} \geq 0\}$, positive $\theta ^{n}$ may be either foreground or background.
Settling this issue requires some extra prior knowledge.
Specifically, we define a $len$ function that counts the features in each set.
For instance, $len(S_{0}) < len(S_{1})$ means that the area of $S_0$ is smaller than $S_1$.
The larger area usually includes image boundary, scene information, or inconspicuous objects and thus is treated as background.
In this way, a $sign$ function is proposed to inverse the activation values if the area of $S_1$ is larger than $S_0$:
\begin{equation}
	sign(\boldsymbol F) = \begin{cases}
	1, &\text{if} \  len(S_{0}) \geq len(S_{1});\\
	-1, &\text{if} \  len(S_{0}) < len(S_{1}).
		   \end{cases}
\end{equation}
Using the $sign$ function, foreground pixels are correspond to positive $\theta ^{n}$.
Secondly, the boundary of salient regions are coarse because of no pixel-level supervision for training.
Therefore, we employ denseCRF ($CRF$) \cite{densecrf} to refine the boundary of foreground regions and Median Filter ($MF$) to remove outliers.
The whole post-process is:
\begin{equation}
\boldsymbol G = MF(CRF(\sigma(sign(\boldsymbol F) * \boldsymbol \Theta )))),
\end{equation}
where $\boldsymbol \Theta = \{\theta^{n}\}$ is the decision value map for each image.
$\boldsymbol G$ is the final pseudo label in the first stage, which will be utilized to train our saliency detector in the next stage.

\subsection{Stage 2: Self-rectification Learning}
\textbf{Saliency Detector.}
Training on pseudo labels is prone to degrading the generalization ability of networks.
Therefore, we develop a simple yet effective saliency detector to better integrate hierarchical representations and refine the learned saliency information.

\begin{figure}[!t]
\centering
\includegraphics[width=3.3in]{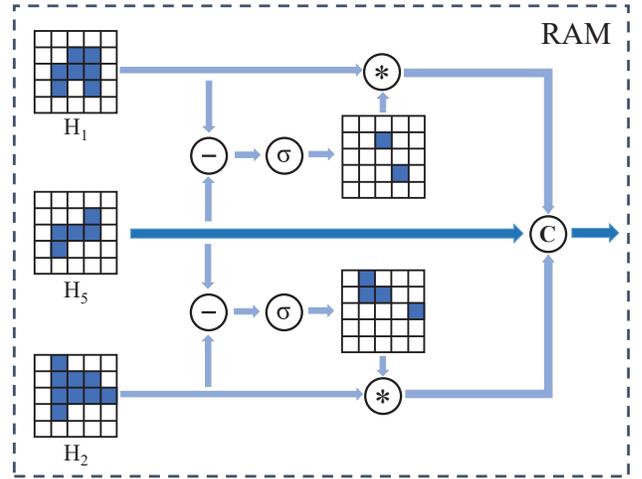}
\caption{Internal structure of our RAM. $-, \sigma, *$ and $C$ indicate subtraction, sigmoid, dot product and concatenation, respectively. $\boldsymbol H_1$ and $\boldsymbol H_2$ can be replaced by $\boldsymbol H_3$ and $\boldsymbol H_4$, respectively.}
\label{fig:ram}
\end{figure}

In our saliency detector, the same pre-trained network is employed as encoder.
In the first stage, we freeze the encoder to preserve the learned semantic information.
In the second stage, we have pixel-level pseudo labels to train our saliency detector.
The fine-tuned encoder can produce more precise predictions.

To reduce the memory footprint, we fix all feature channels to 64 using several convolutional blocks:
\begin{align}
   \boldsymbol E_1, \boldsymbol E_2, \boldsymbol E_3, \boldsymbol E_4, \boldsymbol E_5 &= Encoder_{2}(\boldsymbol X),\\
   \boldsymbol H_i &= conv_i(\boldsymbol E_i), i=1,2,3,4,5.
\end{align}
After that, the topmost feature map $\boldsymbol H_5$ is selected as the basic feature map because its global information can distinguish coarse semantic regions.
The rest four feature maps are employed as supplemental information to refine $\boldsymbol H_5$ from different scales because they contain low-level cues for some other important details.
Specifically, $\boldsymbol H_1$, $\boldsymbol H_2$, $\boldsymbol H_3$ and $\boldsymbol H_4$ are evenly divided into local ($\boldsymbol H_1$ and $\boldsymbol H_2$) and regional ($\boldsymbol H_3$ and $\boldsymbol H_4$) groups according to their receptive fields.
Each group will be combined with $\boldsymbol H_5$ via our Residual Attention Module (RAM), as shown in Fig. \ref{fig:ram}.
Our RAM produces the enhanced feature $\boldsymbol R$ by combining $\boldsymbol H_5$ and two other features $\boldsymbol H_i$.
Notice that all features are upsampled to the same size as the largest input.
In order to learn complementary features, we extract the low-level cues in $\boldsymbol H_i$ by subtracting $\boldsymbol H_5$ from it.
After that, a convolutional layer and a sigmoid function are used to fuse the extracted low-level cues.
Using these cues as attention maps, we strengthen low-level information in $\boldsymbol H_i$ by:
\begin{equation}
   \boldsymbol H'_i = \boldsymbol H_i \odot \sigma(conv(\boldsymbol H_i - \boldsymbol H_5)),,
\end{equation}
where $\odot$ means the Hadamard Product operation.
We then integrate the basic feature map with the enhanced features:
\begin{equation}
\begin{split}
   \boldsymbol R_l &= conv(concat(\boldsymbol H'_1, \boldsymbol H'_2, \boldsymbol H_5)), \\
   \boldsymbol R_r &= conv(concat(\boldsymbol H'_3, \boldsymbol H'_4, \boldsymbol H_5)), \\
\end{split}
\end{equation}
$\boldsymbol R_l$ and $\boldsymbol R_r$ are produced by using local ($\boldsymbol H_1$ and $\boldsymbol H_2$) and regional ($\boldsymbol H_3$ and $\boldsymbol H_4$) groups to enhance $\boldsymbol H_5$, respectively.
Given two outputs, $\boldsymbol R_l$ and $\boldsymbol R_r$, we generate final saliency map $\boldsymbol Y$ with the following:
\begin{equation}
   \boldsymbol Y = sum(\boldsymbol R_l \odot \boldsymbol R_r),
\end{equation}
where $sum$ means aggregating values along the channel dimension.
Finally, $\boldsymbol Y$ is trained by our pseudo label $\boldsymbol G$ with the following loss function:
\begin{equation}
    L_{det} = L_{B}(\boldsymbol Y, \boldsymbol G) + L_{S}(\boldsymbol Y, \boldsymbol G) + L_{I}(\boldsymbol Y, \boldsymbol G),
\end{equation}
where $L_{B}$,  $L_{S}$ and $L_{I}$ are Binary Cross Entropy (BCE), Structural SIMilarity (SSIM) and Intersection-over-Union (IOU) losses \cite{basnet}, respectively.

\begin{figure}[t]
\includegraphics[width=3.3in,height=2.0in]{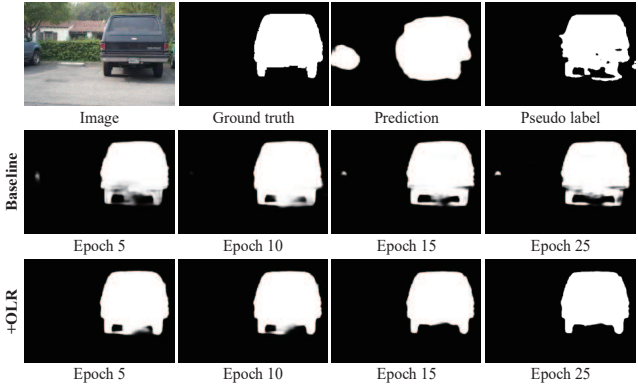}
\caption{ Visualization of the learned saliency maps. Training with our OLR removes some distractors in pseudo labels.}
\label{fig:olr}
\end{figure}

\textbf{Online Label Rectifying.}
We observe that some distractors in pseudo labels may impose a negative impact on training robust saliency detectors.
A visual illustration of our observation is shown in Fig. \ref{fig:olr}.
In the generated pseudo label, the shadow is considered as salient region because its color is similar to wheels.
The detector trained on such labels can distinguish wheels and shadow in early epochs (see 1st image in 2nd row).
However, it produces many low confidence predictions for boundary pixels, resulting in low ave-$F_\beta$ scores.
Moreover, the detector fails to segment both wheels and shadow when more training epochs are done (see 4th image in 2nd row).
To address this issue, an Online Label Rectifying (OLR) strategy is developed to update pseudo labels using the current saliency predictions:
\begin{align}
   \boldsymbol Y^j &= \phi_{det}^{j}(\boldsymbol X),\\
   \boldsymbol G^{j+1} &= \lambda \boldsymbol G^{j} + (1 - \lambda) \boldsymbol Y^j,
   \label{eqn:lambda}
\end{align}
where $j$ means the current training epoch.
$\phi_{det}$ groups all operations in our saliency detector.
Each $\boldsymbol Y^j$ is supervised by corresponding $\boldsymbol G^j$ over the training process.
We initialize $\boldsymbol G^1 = \boldsymbol G$.
$\lambda$ is set to 1 for the first two epochs to prevent pseudo labels from being contaminated by low-quality predictions.
After that, $\lambda$ is set to 0.4 for the rest of training epochs.

\begin{table*}
\caption{Comparison with state-of-the-art SOD methods.
To fairly compared with SBF \cite{sbf}, MNL \cite{mnl} and USPS \cite{usps}, we also report results obtained by the network pre-trained with human annotations (original ImageNet \cite{imagenet} labels), denoted as $\ddagger$.
$\ast$ results are from their papers.
Best scores for both supervised and unsupervised methods are in bold.}
\label{tab:result}
\centering
\renewcommand\arraystretch{1.2}
\renewcommand\tabcolsep{5pt}
\begin{tabular}{l|cc|cc|cc|cc|cc|cc}
\hline
\multirow{2}{*}{Methods} & \multicolumn{2}{c|}{ECSSD}  & \multicolumn{2}{c|}{MSRA-B} & \multicolumn{2}{c|}{DUT-O} & \multicolumn{2}{c|}{PASCAL-S}& \multicolumn{2}{c|}{DUTS-TE}& \multicolumn{2}{c}{HKU-IS} \\\cline{2-13}
& ave-$F_{\beta} \uparrow$ & MAE $\downarrow$ & ave-$F_{\beta} \uparrow$ & MAE $\downarrow$ & ave-$F_{\beta} \uparrow$ & MAE $\downarrow$ & ave-$F_{\beta} \uparrow$ & MAE $\downarrow$ & ave-$F_{\beta} \uparrow$ & MAE $\downarrow$ & ave-$F_{\beta} \uparrow$ & MAE $\downarrow$ \\\hline
\multicolumn{13}{c}{Supervised} \\ \hline
PiCANet \cite{picanet} & 0.885 & 0.046 & 0.874 & 0.055 & 0.710 & 0.068 & 0.804 & 0.076 & 0.759 & 0.041 & 0.870 & 0.044 \\
CPD \cite{cpd} & 0.917 & 0.037 & 0.899 & 0.038 & 0.747 & 0.056 & 0.831 & 0.072 & 0.805 & 0.043 & 0.891 & 0.034 \\
BASNet \cite{basnet} & 0.879 & 0.037 & 0.896 & \textbf{0.036} & 0.756 & 0.056 & 0.781 & 0.077 & 0.791 & 0.048 & 0.898 & 0.033 \\
ITSD \cite{itsd} & 0.906 & 0.036 & 0.902 & 0.038 & \textbf{0.758} & 0.058 & 0.817 & 0.066 & 0.794 & 0.040 & 0.899 & 0.030 \\
MINet \cite{minet} & \textbf{0.924} & \textbf{0.033} & \textbf{0.903} & 0.038 & 0.756 & \textbf{0.055} & \textbf{0.842} & \textbf{0.064} & \textbf{0.828} & \textbf{0.037} & \textbf{0.908} & \textbf{0.028} \\ \hline
\multicolumn{13}{c}{Unsupervised} \\ \hline
DSR \cite{dsr} & 0.639 & 0.174 & 0.723 & 0.121 & 0.558 & 0.137 & 0.579 & 0.260 & 0.512 & 0.148 & 0.675 & 0.143 \\
MC$^\ast$ \cite{mc} & 0.611 & 0.204 & 0.717 & 0.144 & 0.529 & 0.186 & 0.574 & 0.272 & -- & -- & -- & -- \\
RBD \cite{rbd} & 0.686 & 0.189 & 0.751 & 0.117 & 0.510 & 0.201 & 0.609 & 0.223 & 0.508 & 0.194 & 0.657 & 0.178 \\
HS \cite{cssd} & 0.623 & 0.228 & 0.713 & 0.161 & 0.521 & 0.227 & 0.595 & 0.286 & 0.460 & 0.258 & 0.623 & 0.223 \\ \hline
Stage 1 (Ours) & 0.840 & \textbf{0.072} & 0.857 & 0.053 & 0.634 & 0.096 & 0.750 & \textbf{0.114} & 0.693 & 0.080 & 0.822 & 0.056     \\
Stage 1$^\ddagger$ (Ours) & \textbf{0.851} & 0.081 & \textbf{0.881} & \textbf{0.050} & \textbf{0.690} & \textbf{0.080} & \textbf{0.753} & 0.124 & \textbf{0.733} & \textbf{0.073} & \textbf{0.859} & \textbf{0.054} \\ \hline\hline
SBF$^\ddagger$ \cite{sbf} & 0.787 & 0.085 & 0.867 & 0.058 & 0.583 & 0.135 & 0.680 & 0.141 & 0.627 & 0.105 & 0.805 & 0.074 \\
MNL$^{\ddagger\ast}$ \cite{mnl} & 0.878 & 0.070 & 0.877 & 0.056 & 0.716 & 0.086 & \textbf{0.842} & 0.139 & -- & -- & -- & -- \\
USPS$^{\ddagger\ast}$ \cite{usps} & 0.882 & 0.064 & 0.901 & 0.042 & 0.696 & 0.077 & -- & -- & -- & -- & -- & -- \\ \hline
Stage 2 (Ours) & 0.880 & \textbf{0.060} & 0.886 & 0.043 & 0.683 & 0.082 & 0.775 & \textbf{0.106} & 0.714 & 0.076 & 0.856 & 0.048 \\
Stage 2$^\ddagger$ (Ours) & \textbf{0.888} & 0.064 & \textbf{0.902} & \textbf{0.040} & \textbf{0.719} & \textbf{0.069} & \textbf{0.790} & \textbf{0.106} & \textbf{0.750} & \textbf{0.065} & \textbf{0.887} & \textbf{0.042} \\ \hline
\end{tabular}
\end{table*}

\section{Experiment}
\subsection{Experiment settings}
\textbf{Setup.}
All experiments are implemented on a single GTX 1080 Ti GPU using PyTorch \cite{Pytorch}.
Unsupervised ResNet-50 \cite{mocov2} is employed as the encoder in both stages.
Note that no annotated labels are involved here.
The batch size is 8, and images are resized to $320 \times 320$.
For data augmentation, horizontal flipping is employed during the training process.
SGD is utilized as the optimizer of our framework for both stages.
The first stage includes 20 epochs in total with an initial learning rate of 1.
The learning rate is decayed by a factor of 0.1 at 10-th and 16-th epochs.
The second stage includes 25 epochs in total with an initial learning rate of 0.005.
The learning rate is decayed by a factor of 0.1 at 15-th and 20-th epochs.
Our framework takes about 1 hour to complete all training processes on 3000 images.

\textbf{Metrics.}
We utilize ave-$F_{\beta}$ score and Mean Absolute Error (MAE) as criteria.
The formula of $F_{\beta}$ is:
\begin{equation}
F_{\beta} = \frac{(1+\beta^{2})\times Precision \times Recall}{\beta^{2}\times Precision + Recall},
\end{equation}
where $\beta^{2}$ is set to 0.3 \cite{THUR15K} in general.
The ave-$F_{\beta}$ is average over a set of $F_{\beta}$ scores calculated by changing positive thresholds from 0 to 255.
In addition, MAE is obtained by:
\begin{equation}
    MAE = \frac{1}{N}\sum_{n=1}^{N} |  y^{n} - gt^{n}|,
\end{equation}
where $y^n$ and $gt^n$ are $n$-th pixel in prediction and ground truth, respectively.

\textbf{Datasets.}
Following previous DL-based USOD works \cite{usps, mnl}, 3000 images in the train and val subsets of MSRA-B \cite{msra} dataset are employed to train our framework.
ECSSD \cite{ecssd}, PASCAL-S \cite{pascal-S}, HKU-IS \cite{hku-is}, DUTS-TE \cite{duts}, DUT-O \cite{DUT-OMRON} as well as the test subset of MSRA-B are employed as the test sets.
They have 1000, 850, 4447, 5019, 5168, and 2000 images, respectively.

\subsection{Main Results}
We compare our framework with existing state-of-the-art DL-based USOD methods (SBF \cite{sbf}, MNL \cite{mnl} and USPS \cite{usps}) and several traditional methods (DSR \cite{dsr}, MC \cite{mc}, RBD \cite{rbd} and HS \cite{hs}).
Some supervised SOD methods are included, such as PiCANet \cite{picanet}, CPD \cite{cpd}, BASNet \cite{basnet}, ITSD \cite{itsd} and MINet \cite{minet}.
All results are listed in Tab. \ref{tab:result}.

Based on this table, we have four conclusions.
First, the saliency cues extracted in the first stage are significantly more accurate than traditional SOD methods.
Second, our A2S framework achieves competitive results against existing DL-based USOD methods.
It is noteworthy that our framework uses unsupervised encoder, while existing methods take advantage of supervised encoder.
To make a fair comparison, we also report results obtained by the network pre-trained with human annotations (\emph{i.e.}, original ImageNet labels).
As shown in Tab. \ref{tab:result}, our framework achieves state-of-the-art performance compared to all existing DL-based USOD methods.
Third, MNL \cite{mnl} reports a higher ave-F$_\beta$ score on the PASCAL-S dataset, while its MAE score is significantly lower than our method.
Moreover, our framework achieves better performance than MNL on all other test sets.
Last, our framework is competitive to many latest supervised learning based methods on some datasets, such as MSRA-B.

\begin{table}[t]
\begin{center}
\caption{A detailed comparison among DL-based USOD methods.
We collect the train time for each method from its paper.
Note that all existing methods exclude the time for extracting saliency cues.}
\renewcommand\arraystretch{1.2}
\label{tab:compare}
\begin{tabular}{c|ccccc}
\hline
Method & Input & Backbone  & Saliency cues & Train time \\ \hline
SBF \cite{sbf} & $224^2$ & VGG-16  & \cite{mb+, bms, cssd} & $\textgreater$3h \\
MNL \cite{mnl} & $425^2$ & ResNet-101 & \cite{rbd, dsr, mc, cssd} & $\textgreater$4h \\
USPS \cite{usps} & $432^2$ & ResNet-101 & \cite{rbd, dsr, mc, cssd} & $\textgreater$30h \\\hline
Ours & $320^2$ & ResNet-50 & -- & 1h \\
\hline
\end{tabular}
\end{center}
\vspace{-0.15in}
\end{table}

Implementation details in Tab. \ref{tab:compare} further demonstrate the effectiveness and efficiency of our framework.
Our framework use ResNet-50 backbone with $320\times 320$ input, while MNL \cite{mnl} and USPS \cite{usps} use ResNet-101 backbone with larger input.
Moreover, our A2S framework proposes a new method to extract saliency cues based on CNN features, while existing DL-based USOD methods rely on multiple saliency cues from traditional SOD methods.
To compare the efficiency of these methods, we collect the time consumptions from their papers.
SBF \cite{sbf}, MNL \cite{mnl} and USPS \cite{usps} take 3, 4 and 30 hours to finish their training processes.
It is noted that all these methods exclude the time for extracting saliency cues.
Our framework only takes about 1 hour for the whole process, including about 40 and 20 minutes for the first and second stage, respectively.
This comparison well proves that our framework is effective and much more efficient than previous USOD methods.

\begin{figure*}[!t]
\centering
\includegraphics[width=7.2in,height=4.4in]{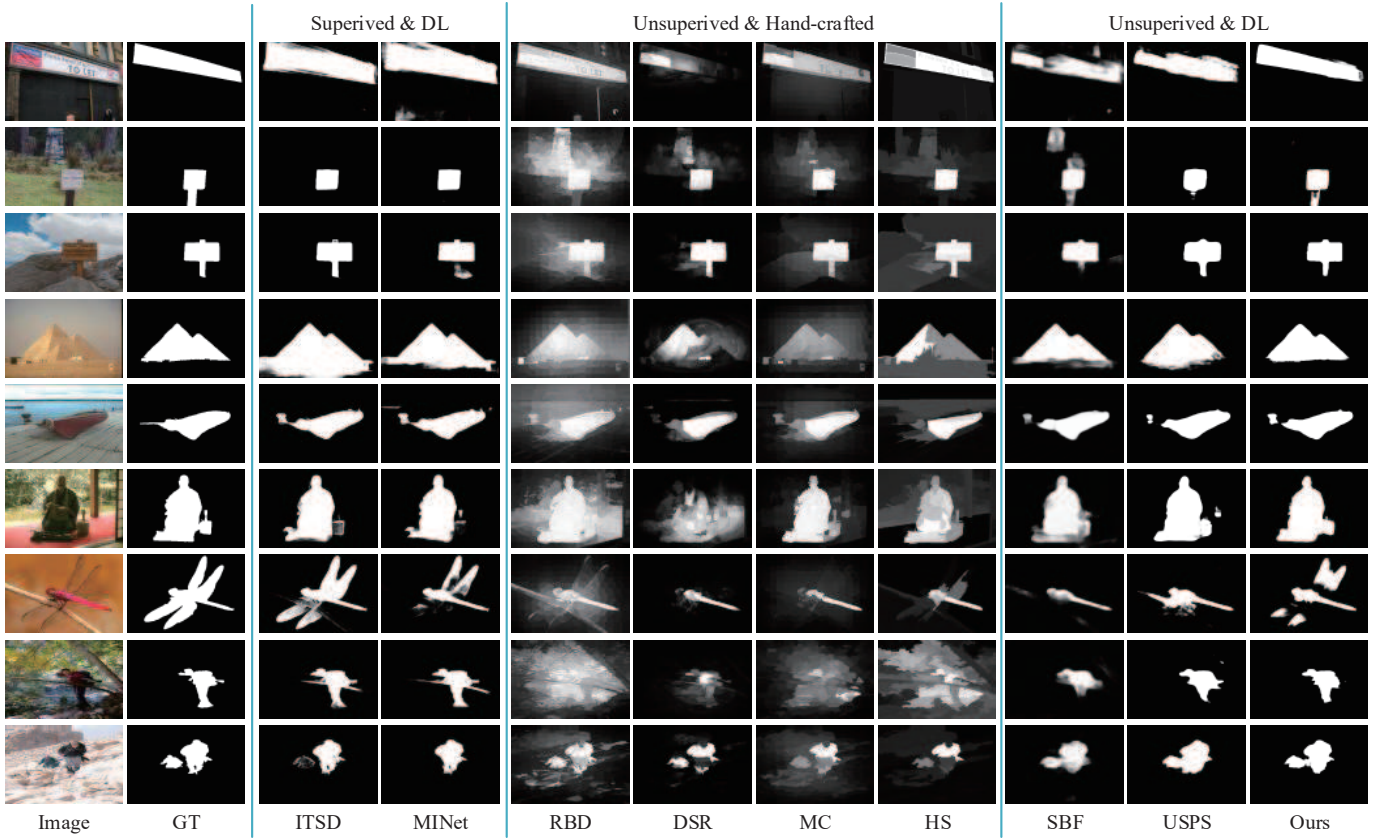}
\caption{Visual comparison with state-of-the-art SOD methods. Predictions of MNL \cite{mnl} has not been published.}
\label{fig:visual}
\end{figure*}

As shown in Fig. \ref{fig:visual}, compared to existing DL-based USOD methods, our method can better distinguish between salient objects and their surrounding backgrounds.
For example, our predictions of the first and fourth images have more distinctive boundaries than previous methods \cite{sbf, usps}.
Compared to traditional SOD methods, our method has significant improvements on the quality of saliency predictions.
Moreover, our method shows competitive performance on all examples compared to supervised DL-based SOD methods.

\begin{table}[t]
\begin{center}
\caption{Ablation study on stage 1. We list the results for different variants of the network in the first stage. All scores are tested on train and val splits of MSRA-B dataset, which are adopted as the train set in our experiment.}
\renewcommand\arraystretch{1.2}
\label{tab:ablation}
\begin{tabular}{c|c|cc}
\hline
Variants & Description  & ave-F$_\beta$ $\uparrow$& MAE $\downarrow$ \\\hline
A0 & Unfreezed encoder & 0.028 & 0.350  \\
A1 & No auxiliary supervision & 0.814 & 0.092  \\
A2 & No dropped pixels & 0.837 & 0.081  \\
A3 & Ours & \textbf{0.866} & \textbf{0.050}  \\\hline
\end{tabular}
\end{center}
\vspace{-0.1in}
\end{table}

\subsection{Ablation Study on Stage 1}
We conduct some experiments to validate the effectiveness of three variant methods of our stage 1.
These variants are: 1) A0: unfreezed encoder; 2) A1: no auxiliary supervision; 3) A2: no pixel dropout in loss function.
We denote our full method as A3.
Results are listed in Tab. \ref{tab:ablation}.

In summary, A3 reports the best results among these variants.
A0 fails to converge because the unsupervised training process destroys the learned semantic information in the encoder.
A1 only supervises the fused feature $F$, and thus the network is hard to conclude more distinctive representations from images.
Since we do not have precise labels for training, A2 using all pixels for training greatly increases the negative impact of overfitting.

\begin{table}[t]
\begin{center}
\caption{Ablation study on $\alpha$. Scores of the generated pseudo labels on the train set are listed.}
\renewcommand\arraystretch{1.2}
\label{tab:alpha}
\begin{tabular}{c|cccc}
\hline

Metric & 0 & 0.1 & 0.2 & 0.3 \\\hline
ave-F$_\beta$ $\uparrow$ & 0.845 & \textbf{0.866} & 0.862 & 0.859 \\
MAE $\downarrow$ & 0.078 & \textbf{0.050} & 0.052 & 0.055 \\\hline
\end{tabular}
\end{center}
\end{table}

\begin{table}[t]
\begin{center}
\caption{Ablation study on $\lambda$. Scores on ECSSD and test split of MSRA-B datasets are listed.}
\renewcommand\arraystretch{1.2}
\label{tab:lambda}
\begin{tabular}{c|c|ccccc}
\hline
Dataset & Metric & 0.2 & 0.3 & 0.4 & 0.5 & 0.6 \\\hline
\multirow{2}{*}{ECSSD} & ave-F$_\beta$ & 0.870 & 0.876 & \textbf{0.880} & 0.879 & 0.873 \\
 & MAE & 0.066 & 0.063 & \textbf{0.060} & 0.062 & 0.065 \\\hline
\multirow{2}{*}{MSRA-B} & ave-F$_\beta$ & 0.878 & 0.882 & \textbf{0.886} & 0.884 & 0.880 \\
 & MAE & 0.048 & 0.046 & \textbf{0.043} & 0.045 & 0.048 \\\hline
\end{tabular}
\end{center}
\end{table}

\subsection{Sensitivity of Hyperparameters}
In this section, we conduct a series of experiments to see the effect of $\alpha$ and $\lambda$ in Eqn. \ref{eqn:loss_1} and \ref{eqn:lambda}, respectively.

\begin{figure}[!t]
\centering
\includegraphics[width=3.1in,height=2.25in]{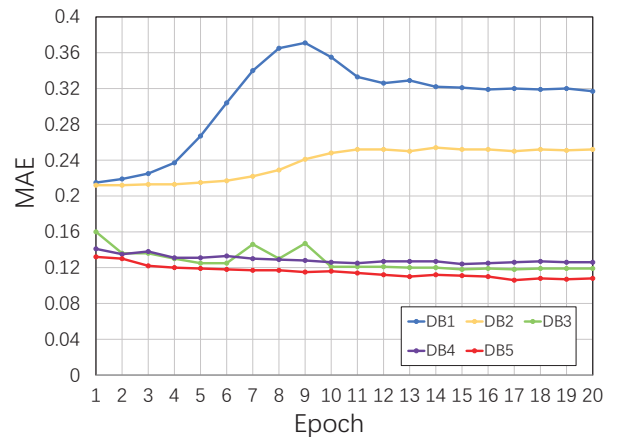}
\caption{MAE curves of different decision boundaries.}
\vspace{-0.1cm}
\label{fig:adb_curve}
\end{figure}

Hyperparameter $\alpha$ in Eqn. \ref{eqn:loss_1} is introduced to adjust the gradients for samples which are closed to the decision boundary.
However, a large $\alpha$ may induce the network pays too much attention on these pixels and ignores other easy samples.
We train the network in the first stage by setting $\alpha$ to 0, 0.1, 0.2, 0.3.
As the results shown in Tab. \ref{tab:alpha}, $\alpha=0$ reports the worst results because gradients for hard samples are vanished.
$\alpha=0.1$ obtains the best results among all competitors.
When $\alpha = 0.2$, the network pays more attention on hard samples, resulting in weakened performance.
As we increase $\alpha$ to 0.3, the network has been witnessed a large performance drop.

Hyperparameter $\lambda$ in Eqn. \ref{eqn:lambda} is designed to control the update process of pseudo labels.
The results of different $\lambda$ are shown in Tab. \ref{tab:lambda}.
Large $\lambda$ means a relatively slow updating rate for pseudo labels.
Distractors in pseudo labels cannot be quickly eliminated, and thus cause inferior results.
Meanwhile, small $\lambda$ means a fast updating rate for pseudo labels.
The networks are supervised by previous estimations, and fails to capture more saliency information from original pseudo labels.

\begin{table}[t]
\begin{center}
\caption{Results of MINet and our saliency detector when training with the generated pseudo labels. $\dagger$ means the proposed RAM is replaced by a convolutional layer.}
\renewcommand\arraystretch{1.2}
\label{tab:minet}
\begin{tabular}{c|c|cc|cc}
\hline
\multirow{2}{*}{Network}  & \multirow{2}{*}{\# Paras. (M)}&  \multicolumn{2}{c|}{ECSSD} & \multicolumn{2}{c}{MSRA-B}  \\ \cline{3-6}
&& ave-$F_{\beta} \uparrow$ & MAE $\downarrow$ & ave-$F_{\beta} \uparrow$ & MAE $\downarrow$ \\ \hline
\multicolumn{6}{c}{w/o OLR}  \\ \hline
MINet & 164.43 & 0.859 & 0.067 & 0.867 & 0.050 \\
Ours$^\dagger$ & 28.25 & 0.860 & 0.068 & 0.865 & 0.051  \\
Ours & 28.40 & 0.865 & 0.066 & 0.872 & 0.047  \\
\hline
\multicolumn{6}{c}{w/ OLR}  \\ \hline
MINet  & 164.43 & 0.863 & 0.065 & 0.872 & 0.048 \\
Ours$^\dagger$  & 28.25 & 0.872 & 0.063 & 0.876 & 0.047 \\
Ours  & 28.40 & 0.880 & 0.060 & 0.886 & 0.043 \\
\hline
\end{tabular}
\end{center}
\end{table}

\begin{figure}[t]
\centering
\begin{minipage}{1 \textwidth}
\includegraphics[width=0.54in,height=0.42in]{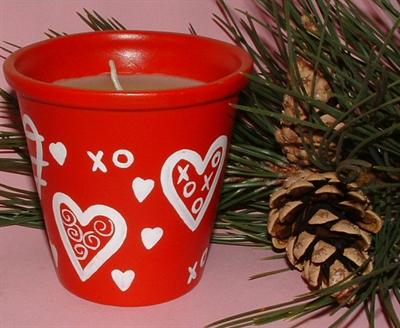}
\includegraphics[width=0.54in,height=0.42in]{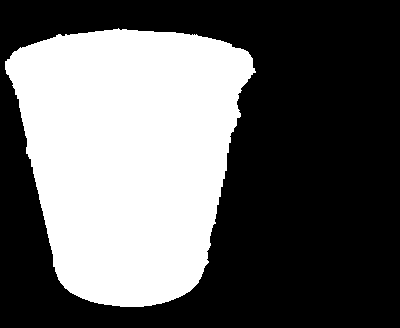}
\includegraphics[width=0.54in,height=0.42in]{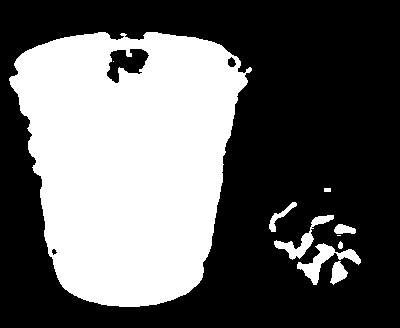}
\includegraphics[width=0.54in,height=0.42in]{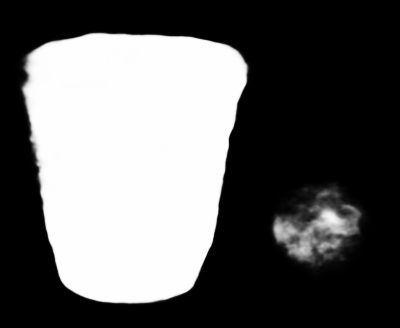}
\includegraphics[width=0.54in,height=0.42in]{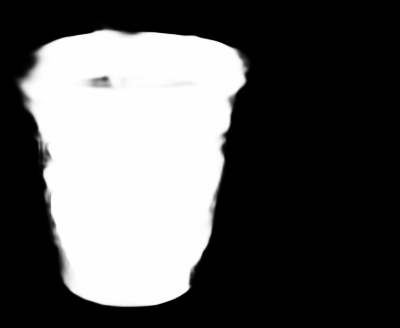}
\includegraphics[width=0.54in,height=0.42in]{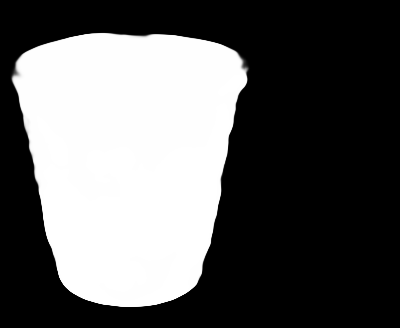}
\end{minipage}

\vspace{0.1cm}
\begin{minipage}{1 \textwidth}
\includegraphics[width=0.54in,height=0.42in]{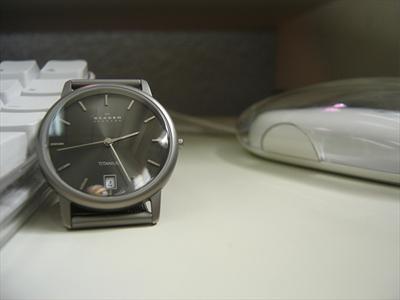}
\includegraphics[width=0.54in,height=0.42in]{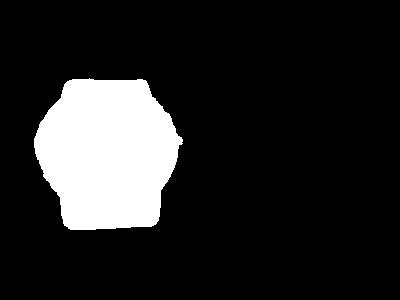}
\includegraphics[width=0.54in,height=0.42in]{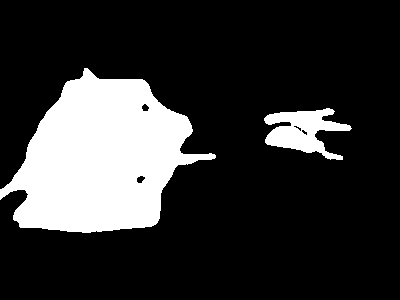}
\includegraphics[width=0.54in,height=0.42in]{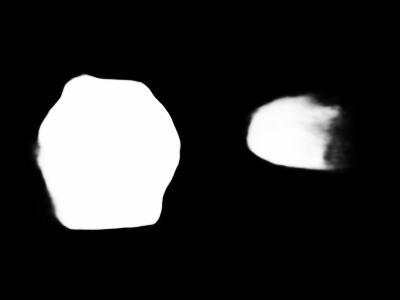}
\includegraphics[width=0.54in,height=0.42in]{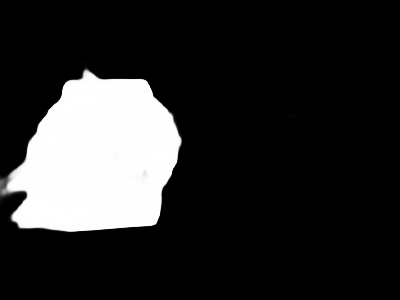}
\includegraphics[width=0.54in,height=0.42in]{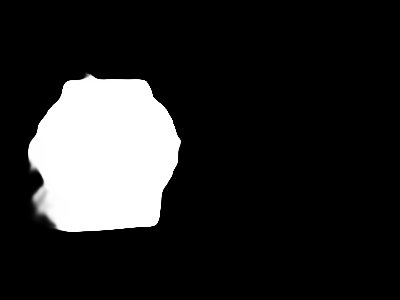}
\end{minipage}

\vspace{0.1cm}
\begin{minipage}{0.99 \textwidth}
\begin{minipage}{0.54in} \centering Image  \end{minipage}
\begin{minipage}{0.54in} \centering GT  \end{minipage}
\begin{minipage}{0.54in} \centering Label  \end{minipage}
\begin{minipage}{0.54in} \centering MINet \end{minipage}
\begin{minipage}{0.54in} \centering Ours$^\dagger$ \end{minipage}
\begin{minipage}{0.54in} \centering Ours \end{minipage}
\end{minipage}

\vspace{0.1cm}
\caption{The learned saliency on training samples. ``Label'' indicates the generated pseudo labels in our first stage. MINet \cite{minet} overfits some distractors in pseudo labels, while our detectors eliminate these regions. }
\label{fig:minet}
\vspace{-0.1cm}
\end{figure}

\subsection{Design of ADB}
In this section, we analyze different designs of our decision boundaries.
We first review and define some symbols used in these decision boundaries:
1) $\boldsymbol{\hat f}$ indicates the mean feature of $\boldsymbol F$ over train set;
2) $\bar{\boldsymbol f}$ indicates the mean feature of $\boldsymbol F$ for each image.
It usually contains information about salient objects;
3) $\boldsymbol w$ means learnable weights (e.g. a fully-connected layer).

Based on these definitions, we define these decision boundaries as:
1) DB1: $\theta_{1}=\textbf{1}^{T}(\boldsymbol f^{n} - \boldsymbol{\hat f})$.
This variant uses a fixed bias term to divide all pixels into two groups;
2) DB2: $\theta_{2}=\bar{\boldsymbol f}^{T}\boldsymbol f^{n}$.
This variant calculates the similarity between $\bar{\boldsymbol f}$ and each feature.
Saliency regions are expected to have higher similarities;
3) DB3: $\theta_{3}=\bar{\boldsymbol f}^{T}(\boldsymbol f^{n} - \bar{\boldsymbol f})$.
The dot product between $\bar{\boldsymbol f}$ and its difference with $\boldsymbol f^{n}$.
4) DB4: $\theta_{4}=\boldsymbol w^{T}(\boldsymbol f^{n} - \bar{\boldsymbol f})$.
This variant uses the learned weights for the difference between $\boldsymbol f^{n}$ and $\bar{\boldsymbol f}$;
5) DB5: $\theta_{5}=\boldsymbol 1^{T}(\boldsymbol f^{n} - \bar{\boldsymbol f})$.
This variant is the proposed ADB in our framework.
Their MAE curves during the training process are shown in Fig. \ref{fig:adb_curve}.

Overall, our ADB (DB5) reports the best and most stable performance among all competitors.
DB1 and DB2 fail to converge due to the unbalanced ratio between two groups.
They are prone to splitting all pixels into the same class.
DB3 is unstable because background pixels greatly disturb the semantic cues in $\bar{\boldsymbol f}$.
Since the network in the first stage is trained by unsupervised learning, our supervision signals are significantly coarse than human annotations.
Therefore, to improve the performance of DB4, more constraints need to be imposed on the learning process of $w$.

\subsection{Our Detector vs. State-of-the-art}
We conduct an experiment to compare our detector with MINet \cite{minet}, which has reported state-of-the-art performance on various SOD benchmarks.
We train these networks with our pseudo labels and report results in Tab. \ref{tab:minet}.

No matter using OLR or not, our detector reports significant improvements compared with MINet.
MINet is likely to be overfitted to some distractors in pseudo labels due to its large amount of trainable parameters, as shown in Fig. \ref{fig:minet}.
The lightweight structure of our detector alleviates the overfitting problem and thus results in better performance.
In addition, RAMs further improve the performance of our saliency detector.

\begin{figure}[!t]
\centering
\begin{minipage}{1 \textwidth}
\includegraphics[width=0.82in,height=0.68in]{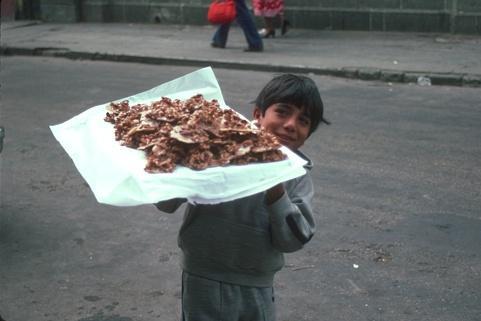}
\includegraphics[width=0.82in,height=0.68in]{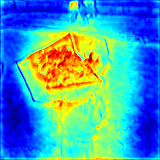}
\includegraphics[width=0.82in,height=0.68in]{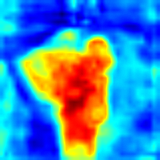}
\includegraphics[width=0.82in,height=0.68in]{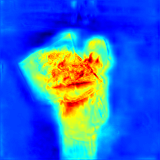}
\end{minipage}

\vspace{0.1cm}
\begin{minipage}{1 \textwidth}
\includegraphics[width=0.82in,height=0.68in]{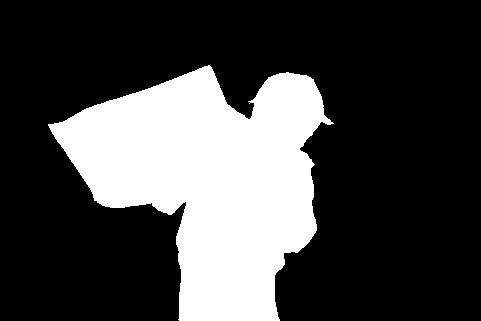}
\includegraphics[width=0.82in,height=0.68in]{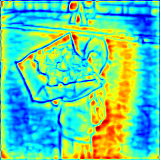}
\includegraphics[width=0.82in,height=0.68in]{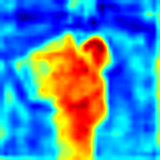}
\includegraphics[width=0.82in,height=0.68in]{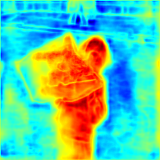}
\end{minipage}

\vspace{0.1cm}
\begin{minipage}{1 \textwidth}
\includegraphics[width=0.82in,height=0.68in]{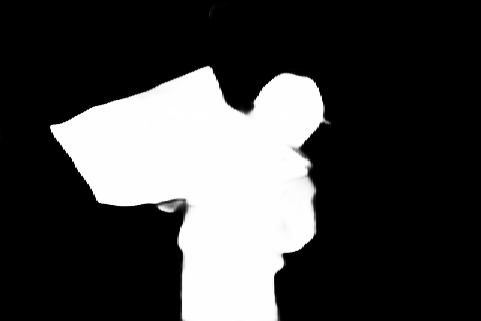}
\includegraphics[width=0.82in,height=0.68in]{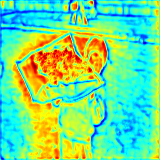}
\includegraphics[width=0.82in,height=0.68in]{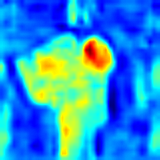}
\includegraphics[width=0.82in,height=0.68in]{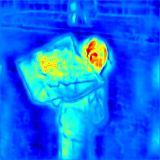}
\end{minipage}

\vspace{0.1cm}
\begin{minipage}{0.99 \textwidth}
\begin{minipage}{0.82in} \centering \quad  \end{minipage}
\begin{minipage}{0.82in} \centering $\boldsymbol R_l$ \end{minipage}
\begin{minipage}{0.82in} \centering $\boldsymbol R_r$ \end{minipage}
\begin{minipage}{0.82in} \centering $\boldsymbol R_l \odot \boldsymbol R_r$ \end{minipage}
\end{minipage}

\vspace{0.1cm}
\caption{Visualization of the learned features in RAMs.
As a reference, we show image, ground truth and final prediction in the first column (from top to bottom).}
\label{fig:fusion}
\end{figure}

\begin{table}[t]
\begin{center}
\caption{Comparison between HMA and our OLR.}
\renewcommand\arraystretch{1.2}
\label{tab:hma}
\begin{tabular}{c|c|cc|cc}
\hline
\multirow{2}{*}{Method}  & \multirow{2}{*}{Tr. time(h)}  & \multicolumn{2}{c|}{ECSSD} & \multicolumn{2}{c}{MSRA-B}  \\ \cline{3-6}
&& ave-$F_{\beta} \uparrow$ & MAE $\downarrow$ & ave-$F_{\beta} \uparrow$ & MAE $\downarrow$ \\ \hline
HMA \cite{usps} & 5.33 & 0.843 & 0.078 & 0.876 & 0.049 \\
OLR (Ours) & 0.36 & 0.880 & 0.060 & 0.886 & 0.043 \\
\hline
\end{tabular}
\end{center}
\end{table}

\subsection{Visualization of RAM}
We visualize the learn features in the proposed RAMs in Fig. \ref{fig:fusion}.
Activations from $\boldsymbol R_l$ contain many low-level cues, such as points and plane edges, while responses from $\boldsymbol R_r$ show more region-wise patterns.
After multiplications, feature maps in $\boldsymbol R_l \odot \boldsymbol R_r$ well integrate multi-level information in $R_l$ and $R_r$, and activate some regions of salient objects.
By summing all feature maps in $\boldsymbol R_l \odot \boldsymbol R_r$, the network aggregates these regions and compose the final predictions.

\subsection{OLR vs. HMA}
Similar to the proposed OLR, the Historical Moving Averages (HMA) strategy was proposed in previous work \cite{usps}.
Our OLR is different with HMA from several perspectives.
First, HMA generates pseudo labels from low-quality saliency cues, while our OLR aims at rectifying the pseudo labels online.
Second, HMA uses CRF to refine the network predictions before updating, while our OLR uses these predictions to eliminate distractors introduced by CRF.
Last, HMA is much slower than our OLR due to the CRF operation.

To prove our points, we use HMA and the proposed OLR to train our detector.
As shown in Tab. \ref{tab:hma}, the proposed OLR achieves significantly improvements compared with HMA.
Moreover, using the proposed OLR is 15$\times$ faster than HMA.

\section{Conclusion}
In this work, we propose a two-stage Activation-to-Saliency (A2S) framework for Unsupervised Salient Object Detection (USOD) task.
In the first stage, we transform a pre-trained network to generate a single activation map from each image.
An Adaptive Decision Boundary (ADB) is proposed to facilitate the generation of high-quality pseudo labels.
In the second stage, we construct a lightweight saliency detector with an encoder and two Residual Attention Modules (RAMs) to alleviate the overfitting problem. An Online Label Rectifying (OLR) strategy is used to update the pseudo labels for reducing the negative impact of distractors.
Extensive experiments on several SOD benchmarks prove the effectiveness and efficiency of the proposed framework.

\bibliographystyle{IEEEtran}
\bibliography{egbib}

%








\end{document}